\documentclass[times, review, 10pt]{elsarticle}

\usepackage{enumitem}
\usepackage{amssymb}
\usepackage{caption}
\usepackage{amsmath}
\usepackage{pifont}
\usepackage{subfigure}
\usepackage{amsfonts}       
\usepackage{nicefrac}       
\usepackage{microtype}      
\usepackage{amsthm}
\usepackage{ulem}
\usepackage[ruled]{algorithm2e}
\usepackage{booktabs}
\usepackage{multirow}
\usepackage{soul}
\usepackage{bm}
\usepackage{color, xcolor}
\usepackage{tikz}
\usepackage{fontawesome}
\definecolor{deepred}{RGB}{192,0,0}
\definecolor{deepyellow}{RGB}{197,90,17}
\definecolor{deepblue}{RGB}{0,128,172}
\definecolor{deepgreen}{RGB}{84,130,53}
\definecolor{skyblue}{RGB}{0,191,255}
\definecolor{indigo}{RGB}{102,0,153}
\definecolor{darkred}{RGB}{0.7, 0, 0} 
\definecolor{darkblue}{RGB}{0.0, 0.0, 0.7} 
\definecolor{myblue}{RGB}{27,183,251}
\usepackage{hyperref}

\hypersetup{
    colorlinks=true,
    linkcolor=deepblue,
    citecolor=deepblue,
    filecolor=deepblue,
    urlcolor=deepblue,
}

\usepackage[table]{xcolor}

\usepackage{makecell}
\begin{document}
\begin{frontmatter}



\title{MAETrack: Unleashing the Potential of Pretrained Geometric Priors for 3D Single Object Tracking}

\author[a,b]{Sifan Zhou$^\dagger$\corref{cor1}}
\author[c]{Qiwei Wang$^\dagger$}
\author[d]{Linyue Tan$^\dagger$}
\author[d]{Ziyu Liu}
\author[a,b]{Ziyu Zhao}
\author[a,b]{Xiaobo Lu\corref{cor1}}

\cortext[cor1]{Corresponding author: Sifan Zhou, sifanjay@gmail.com; Xiaobo Lu, xblu2013@126.com. $^\dagger$ Equal Contribution.}

\affiliation[a]{
    organization={School of Automation, Southeast University},
    city={Nanjing},
    postcode={210096},
    country={China}
}

\affiliation[b]{
    organization={Key Laboratory of Measurement and Control of Complex Systems of Engineering, \\ Ministry of Education},
    city={Nanjing},
    postcode={210096},
    country={China}
}

\affiliation[c]{
    organization={Harbin Institute of Technology Shenzhen},
    city={Shenzhen},
    postcode={518000},
    country={China}
}

\affiliation[d]{
    organization={University of Pennsylvania },
    city={Philadelphia, PA},
    postcode={19104},
    country={USA}
}

\begin{abstract}
Large-scale pre-training has transformed representation learning in 2D vision, yet its transferability to 3D single object tracking (SOT) remains insufficiently understood. Directly fine-tuning self-supervised 3D encoders, such as masked autoencoders (MAE), often leads to sub-optimal adaptation because the reconstruction objective is not fully aligned with the spatial-temporal matching requirements of tracking. In this paper, we observe that this difficulty can be interpreted as a layer-wise transfer mismatch: shallow layers tend to preserve transferable geometric cues, while deeper layers become increasingly specialized to the reconstruction pretext task and are less suitable for downstream tracking. Based on this observation, we propose \textbf{MAETrack}, a lightweight adaptation framework for transferring pre-training MAE representations to 3D SOT. \textbf{MAETrack} includes \textbf{Layer-Selective Initialization (LSI)}, which initializes only the shallow stages of the tracking backbone from pre-trained weights while re-initializing deeper stages, and \textbf{Geometric Residual Gating (GRG)}, which reinforces structurally salient regions in the search BEV features before template-search fusion through residual spatial modulation. Extensive experiments on standard 3D SOT benchmarks show that MAETrack consistently improves upon vanilla fine-tuning baselines with limited computational overhead. More broadly, our results suggest that effective transfer from 3D reconstruction pre-training to 3D tracking is not merely a matter of partial fine-tuning, but depends on a tracking-oriented transfer principle that preserves shallow geometry while adapting deeper representations to the downstream objective. Code will be released after acceptance.
\end{abstract}


\end{frontmatter}

\section{Introduction}\label{sec:intro}
Large-scale self-supervised pre-training has fundamentally reshaped the landscape of 2D computer vision, empowering downstream applications with robust and generalizable representations in computer vision~\cite{lu2024hrnet,shi2025rethinking,zhou2024lidarptq}, and robotics~\cite{Wang_Zhang_Dodgson_2025,cui2026dct,tartanimu}. Inspired by this success, masked autoencoding methods have significantly advanced 3D representation learning by reconstructing masked observations from partial point clouds~\cite{pang2022masked, min2022voxel,khan2024beyond}. Through this reconstruction-based pre-training objective, MAE-style models learn rich geometric priors and strong sensitivity to local topology and structural continuity, achieving strong performance in scene understanding tasks such as 3D object detection~\cite{zhou2023fastpillars,pillarhist}. 

However, transferring such pretrained representations to 3D Single Object Tracking (SOT) remains challenging. Unlike static perception tasks, 3D SOT requires fine-grained, instance-level spatial-temporal reasoning to continuously localize a target under sparse, partial, and occluded observations. Given the tracked target in the first frame of a point cloud sequence, the goal of 3D SOT is to continuously localize the same object in subsequent frames~\cite{m2track++,zhao2025vpmcan}. In practice, the quality of target representations can degrade substantially when the observed geometry becomes weak or partially missing, making robust feature transfer especially important for 3D SOT.or 3D SOT. 

\begin{figure}[!htbp]
\centering
\includegraphics[width=0.9\linewidth]{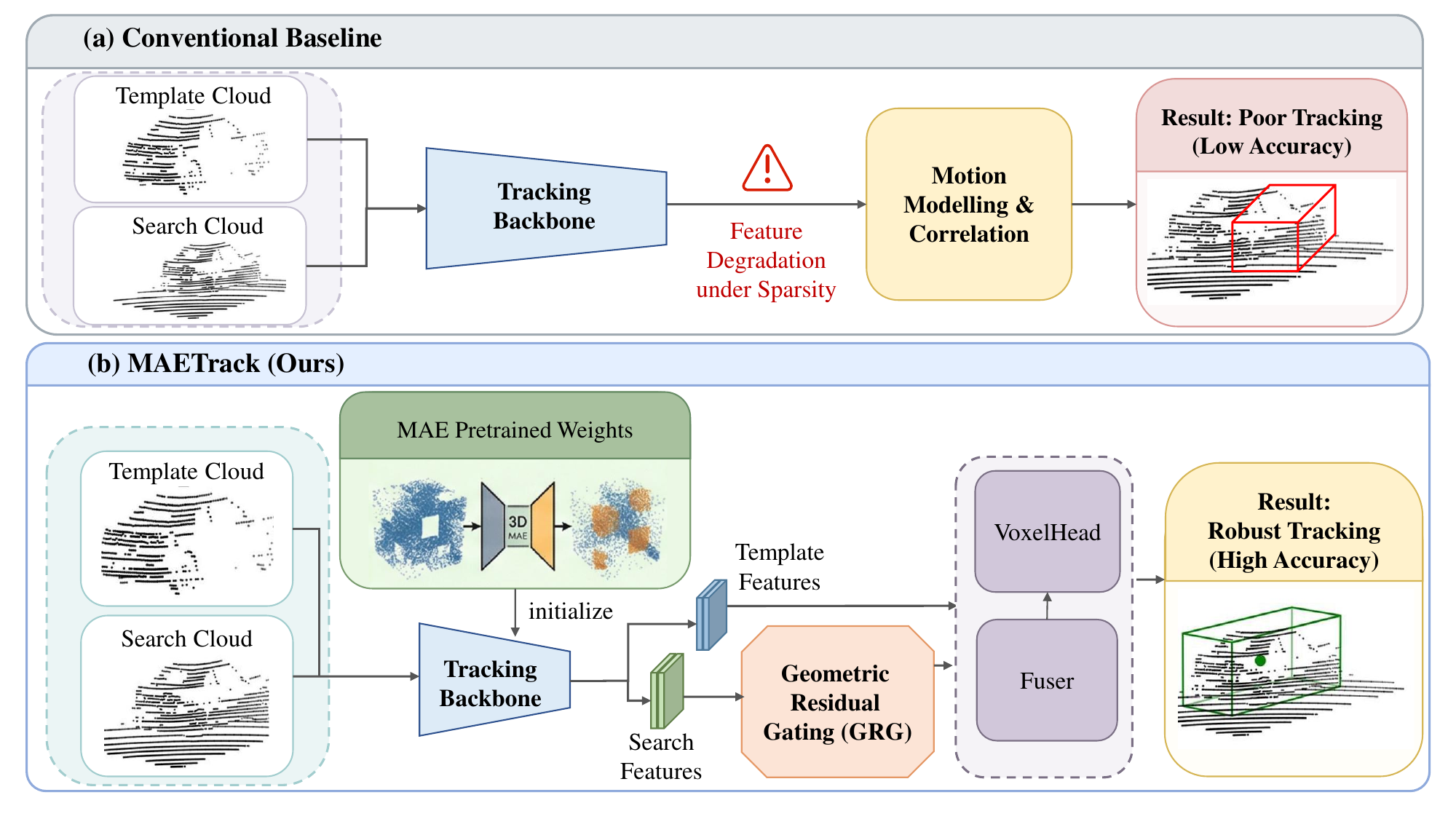}
\caption{\textbf{Comparison between the conventional baseline and our proposed MAETrack.} (a) Conventional trackers suffer from severe feature degradation under point cloud sparsity and ultimately leading to poor tracking accuracy. (b) Our MAETrack overcomes this by initializing the backbone with MAE-pretrained weights for robust feature extraction, and introducing a Geometric Residual Gating (GRG) module to refine the search features, achieving robust tracking with better accuracy and efficiency.}
\label{fig:method-comparison}
\end{figure}

We argue that this difficulty stems from a fundamental layer-wise objective mismatch between the reconstruction-based pre-training and the downstream tracking objective. Masked reconstruction favors dense and position-sensitive geometric recovery, which may cause deeper representations to become increasingly specialized to semantic or structural patterns useful for inferring missing observations. While beneficial for holistic perception, such deep representations may be less compatible with the strict spatial and motion consistency required by tracking-oriented matching. Conversely, the shallow layers of pre-trained models tend to preserve more transferable geometric and boundary cues. This suggests a key but underexplored phenomenon: \textbf{different layers of MAE-pretrained models encode fundamentally different types of geometric information, with shallow layers being more transferable and deep layers being more task-specific.} From this perspective, 3D SOT relies more heavily on shallow geometric priors for accurate localization under sparse or partially occluded observations, whereas deep reconstruction-specialized semantic representations may introduce misalignment with tracking objectives. This layer-wise discrepancy therefore provides a natural motivation for selective transfer of MAE-pretrained representations.

The prevailing paradigm for using pretrained models in 3D SOT is vanilla fine-tuning, which optimizes the downstream tracking objective in an end-to-end manner. However, this direct transfer strategy lacks an explicit mechanism for preserving and adapting pretrained geometric priors to tracking-oriented features. As a result, the geometric knowledge learned during MAE pre-training can be under-utilized during downstream optimization. This limitation becomes more evident in challenging tracking scenarios, such as sparse, ambiguous, or partially occluded target observations, where stronger geometric support may be especially beneficial. More importantly, the core bottleneck is not merely optimization instability, but the mismatch between what different pre-trained layers encode and what downstream tracking actually requires. Therefore, an effective solution should account for this transfer gap rather than rely solely on stronger end-to-end fine-tuning.

To address this issue, we propose \textbf{MAETrack}, a simple yet effective adaptation framework that bridges reconstruction-oriented pre-training and 3D single object tracking. Recognizing the layer-wise objective mismatch between the two tasks, MAETrack introduces a \textbf{Layer-Selective Initialization (LSI)} strategy. Instead of conventional full-network fine-tuning, LSI initializes only the shallow layers of the tracking backbone with BEV-MAE weights to inherit transferable geometric edges and occupancy cues, while randomly initializing the deeper layers to reduce the influence of reconstruction-specific semantics. To further exploit these inherited geometric properties during downstream matching, we design \textbf{Geometric Residual Gating (GRG)}, a lightweight spatial modulation module applied to the search BEV features prior to template-search fusion. Formulated as a self-driven residual scaling mechanism, GRG modulates the search representation based solely on its own spatial context, enabling the tracker to reinforce structurally salient regions without disturbing the main spatial-temporal pipeline. We validate MAETrack on standard 3D single object tracking benchmarks, showing consistent improvements over conventional full-network fine-tuning baselines in both Success and Precision. Our study suggests that the key challenge in transferring MAE-pretrained representations to 3D SOT is not whether geometric priors are useful, but how they should be adapted to the downstream tracking objective. Our contributions can be summarized as:

\begin{itemize}
    \item We identify the layer-wise adaptation gap between reconstruction-based pre-training and downstream tracking, revealing that conventional full-network fine-tuning suffers from potentially mismatched deep representations.
    \item We propose \textbf{MAETrack}, a lightweight 3D SOT framework built upon \textbf{Layer-Selective Initialization (LSI)} and \textbf{Geometric Residual Gating (GRG)}. LSI prevents negative semantic transfer by selectively inheriting shallow geometric weights, while GRG achieves a safe, spatially-aware adaptation of these priors via dynamic residual gating mechanism on the search prior to temporal fusion.
    \item We conduct extensive experiments on standard 3D single object tracking benchmarks and demonstrate that MAETrack consistently improves performance over the strong baseline with negligible overhead.
\end{itemize}

\section{Related Work}\label{sec:related}
\subsection{Point Cloud 3D Single Object Tracking.} 3D SOT on LiDAR point clouds has evolved from early Siamese paradigms to recent motion-centric methods~\cite{lttr,cui2019point,hu2026tftrack}. Early works framed object tracking as a similarity matching problem within a Siamese architecture. For instance, SC3D~\cite{sc3d} utilized a Siamese network to determine the target state through exhaustive feature distance ranking between the template and candidate seeds. To enhance efficiency, 3D-SiamRPN~\cite{3dsiamrpn} introduced the Region Proposal Network (RPN) into the 3D domain, while P2B~\cite{p2b} leveraged a Hough voting strategy to generate high-quality candidates in an end-to-end manner. Subsequent research has sought to refine this paradigm by incorporating sophisticated structural priors and attention mechanisms; for instance, BAT~\cite{bat} integrated box-level size priors, while PTT~\cite{ptt} introduced transformer architectures to capture long-range dependencies. To address the inherent sparsity and occlusion challenges, MBPTrack~\cite{mbptrack} utilized an external memory bank to aggregate multi-frame information. Despite these advancements, appearance matching-based methods remain sensitive to the lack of discriminative textures in LiDAR point clouds. This has catalyzed a shift toward motion-centric paradigms that model inter-frame dynamics. M$^2$Track~\cite{m2track,m2track++} pioneered this direction by simplifying tracking into 4-DOF relative motion regression, and P2P~\cite{p2p} proposed a part-to-part motion modeling framework that serves as a strong baseline for modern 3D SOT. 

However, both Siamese-based matching and motion-based regression currently rely heavily on task-specific supervision from labeled tracking datasets. While self-supervised learning, particularly Masked Autoencoders (MAE)~\cite{he2022masked}, has shown remarkable potential in learning robust, generalized geometric representations for 3D scene understanding, its application within the 3D SOT pipeline remains largely unexplored. Specifically, the community has yet to explicitly investigate how to transfer MAE-pretrained geometric knowledge to enhance tracking robustness against sparsity and occlusion, leaving the gap in leveraging large-scale self-supervised priors for real-time object tracking.

\subsection{3D Vision MAE.} Self-supervised pre-training via Masked Autoencoders (MAE) has revolutionized visual representation learning. Pioneered by MAE~\cite{he2022masked} in the 2D domain, this approach demonstrated that reconstructing masked image patches is a powerful pretext task for learning high-level semantic abstractions with remarkable scalability. However, the direct migration of this paradigm to 3D vision is non-trivial due to the inherent sparsity and unstructured nature of point clouds. To bridge this gap, point- and voxel-based variants such as Point-MAE~\cite{pang2022masked} and Voxel-MAE~\cite{min2022voxel} were developed, which adapt the masking strategy to 3D coordinate spaces or discretized grids. Subsequent research has further diversified the MAE landscape: GeoMAE~\cite{tian2023geomae} incorporates geometric constraints for finer structural modeling, while BEV-MAE~\cite{lin2024bevmae} and Multi-view MAE explore masked reconstruction at the bird’s-eye-view or multi-perspective levels for large-scale outdoor scenes. While these methods have proven effective in static 3D perception tasks—such as object classification and detection~\cite{zhou2023fastpillars}—their utility is predominantly measured by their ability to capture intra-frame geometric integrity.

However, static success does not seamlessly translate to 3D SOT, which demands discriminative cross-frame association over mere geometry reconstruction. The generative nature of MAE-based priors may conflict with tracking’s discriminative requirements, particularly as focus shifts to motion modeling in deeper layers. Consequently, the synergy between self-supervised structural learning and real-time localization remains virtually unexplored in the 3D SOT community.

\begin{figure}
\centering
\includegraphics[width=\linewidth]{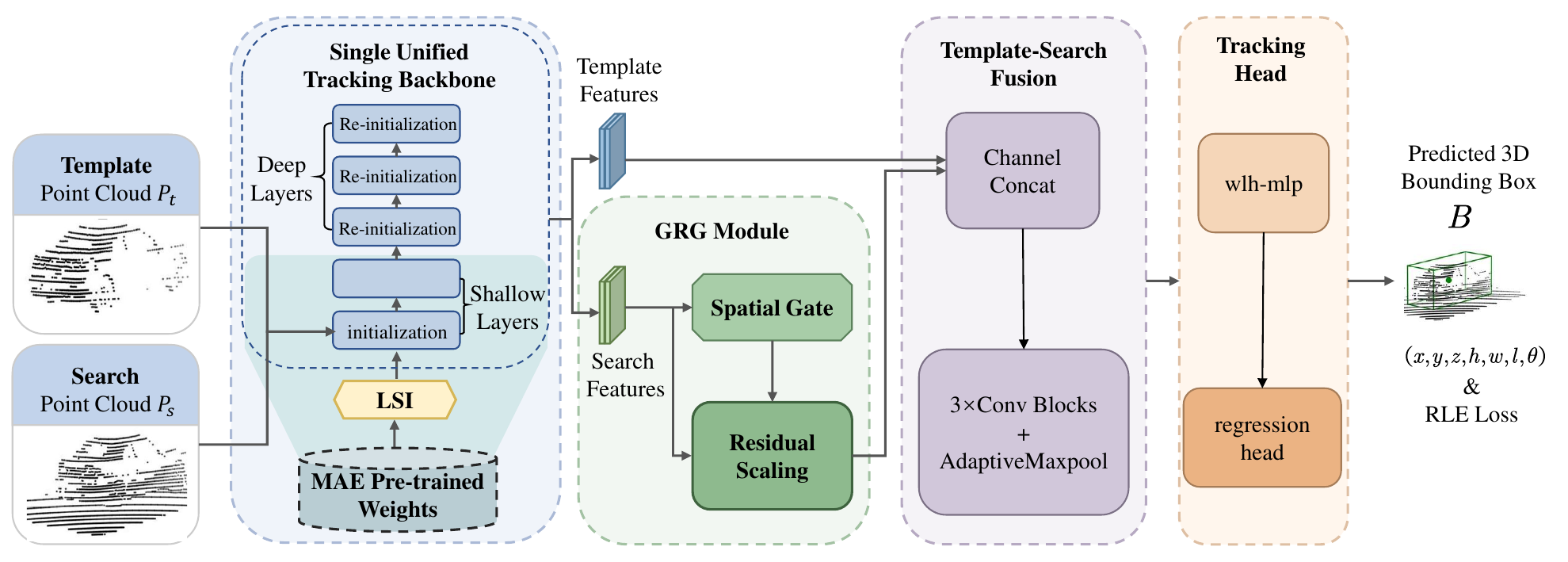}
\caption{Framework of the proposed MAETrack. The pipeline takes a template point cloud and a search point cloud as inputs with an unified feature extraction. To inherit robust geometric priors, its shallow layers are initialized from MAE-pre-trained weights via Layer-Selective Initialization (LSI), while the deep layers are randomly re-initialized to adapt to the tracking task. Then, the search features are fed into the Geometric Residual Gating (GRG) module, which learns a spatial gating map through a spatial gate to selectively enhance informative geometric patterns via residual scaling. Finally, the template and refined search features are aggregated in the template-search fusion module and input the tracking head to predict the 3D bounding box.}
\label{fig:maetrack-overview}
\vspace{-3mm}
\end{figure}

\section{Methodology}\label{sec:method}
\subsection{Problem Formulation} 
Given a sequence of 3D point clouds, the goal of 3D single object tracking (SOT) is to continuously localize a specific target, typically initialized by a 3D bounding box $\mathbf{B}=\{\mathbf{b}=[x, y, z, h, w, l, {\theta}]^T \in \mathbb{R}^{1 \times 7}\}$ in the first frame. Many recent frameworks~\cite{ptt-journal,focustrack} rely on template-search feature extraction and fusion to predict the target state frame by frame. Let $\mathcal{P}_{temp}$ and $\mathcal{P}_{search}$ denote the point clouds of the template (target) and the search region, respectively, where $\mathcal{P}_{temp}=\{\mathbf{p}_{i}^{t}=[x_i, y_i, z_i, r_i]^T \in \mathbb{R}^{N \times 4}\}$, where $x_i, y_i, z_i$ denote the coordinate values of each point. A shared backbone extracts their Bird's-Eye-View (BEV) features, which are subsequently fused for target localization. In our setting, the prediction head mainly regresses the target center and heading angle, while the box size is initialized from the template commonly used in prior 3D SOT pipelines. Formally, the tracking objective can be formulated as:
\begin{equation}
\mathbf{B}_{t}
=
\mathcal{F}_{track}
(
\mathcal{P}_{temp},
\mathcal{P}_{search},
\mathbf{B}_{t-1};
\Theta
),
\end{equation}
where $\mathcal{F}_{track}$ denotes the learnable tracking network parameterized by $\Theta$, and $\mathbf{B}_{t}$ represents the predicted target state at frame $t$.

\subsection{Motivation} 
While pre-trained foundation models~\cite{tian2023geomae} provide strong geometric representations, naively fine-tuning the entire network for 3D SOT often leads to negative transfer. This suggests reconstruction-oriented pre-training and downstream tracking optimize different objectives: the former emphasizes recovering missing geometric content, while the latter requires robust target-specific spatial-temporal matching. 

From a representation learning perspective, this discrepancy can be attributed to a layer-wise specialization phenomenon, where shallow layers tend to preserve local geometric structures such as occupancy patterns and boundaries, while deeper layers encode reconstruction-oriented semantic abstractions~\cite{zhou2023fastpillars,pillartrack}. This leads to a mismatch when directly transferring all layers to tracking tasks, which rely more on instance-level spatial consistency rather than reconstruction fidelity. Based on this observation, we formulate \textbf{MAETrack} as a structured representation adaptation problem, where the goal is not to uniformly fine-tune pretrained features, but to explicitly account for the hierarchical structure of MAE representations. In particular, we aim to selectively preserve transferable geometric components in shallow layers while allowing higher-level representations to adapt to tracking-specific objectives. 

To formulate this transfer problem, we denote a pretrained MAE backbone with $L$ hierarchical stages as:
\begin{equation}
\Theta^{\mathrm{MAE}}
=
\{\theta_1^{\mathrm{MAE}},
\theta_2^{\mathrm{MAE}},
\dots,
\theta_L^{\mathrm{MAE}}\},
\end{equation}
where $\theta_i^{\mathrm{MAE}}$ represents the parameters of the $i$-th stage. Conventional fine-tuning directly initializes all layers from the pretrained model:
\begin{equation}
\Theta^{\mathrm{init}}
=
\Theta^{\mathrm{MAE}}.
\end{equation}
However, this uniform transfer strategy implicitly assumes that all layers share identical transferability, which contradicts the layer-wise specialization property of MAE representations. To characterize the transfer compatibility of each layer, we define:
\begin{equation}
T_i
=
\mathcal{D}
\left(
\theta_i^{\mathrm{MAE}},
\theta_i^{\mathrm{trk}}
\right),
\end{equation}
where $\mathcal{D}(\cdot,\cdot)$ measures the representation compatibility between the pretrained and tracking-oriented parameters at the $i$-th stage. Due to the different optimization objectives, shallow layers generally exhibit higher transfer compatibility than deeper layers:
$T_1,T_2,\dots,T_k > T_{k+1},\dots,T_L.$

Based on this observation, we formulate \textbf{MAETrack} as a structured representation adaptation problem, where the goal is not to uniformly fine-tune pretrained features, but to explicitly account for the hierarchical structure of MAE representations. Specifically, we aim to preserve transferable geometric components in shallow layers while allowing higher-level representations to adapt to tracking-specific objectives.

Accordingly, we introduce a layer-wise initialization operator $\mathcal{A}(\cdot)$ to guide the transfer process:
\begin{equation}
\theta_i^{\mathrm{init}}
=
m_i\theta_i^{\mathrm{MAE}}
+
(1-m_i)\theta_i^{\mathrm{rand}},
\end{equation}
where $m_i$ is a binary transfer mask. Specifically, $m_i=1$ indicates that the corresponding layer inherits MAE-pretrained weights, while $m_i=0$ denotes random initialization. This formulation provides the foundation of our \textbf{Layer-Selective Initialization (LSI)}, which preserves shallow geometric priors while decoupling deeper reconstruction-specific representations.

After initialization, all parameters remain trainable during downstream optimization:
\begin{equation}
\Theta^{\mathrm{trk}}
=
\Theta^{\mathrm{init}}
-
\eta
\nabla_{\Theta}
\mathcal{L}_{\mathrm{track}},
\end{equation}
where $\eta$ denotes the optimization learning rate and $\mathcal{L}_{\mathrm{track}}$ represents the tracking objective. Therefore, LSI does not constrain the final parameter space, but instead provides a geometry-aware initialization trajectory for adapting pretrained representations.

We further decompose this layer-wise representation mismatch into two coupled factors that jointly determine the effectiveness of MAE transfer for 3D SOT. \textit{First}, shallow layers encode transferable geometric structures that should be preserved during adaptation, while deep layers encode reconstruction-specific semantics that may hinder tracking performance. This motivates a selective transfer strategy that preserves shallow geometric representations while decoupling deep semantic components, leading to \textbf{Layer-Selective Initialization (LSI)}. \textit{Second}, even within transferable shallow representations, not all spatial regions contribute equally to tracking. In particular, structurally salient regions such as boundaries and occupied areas are more informative under sparse observations. This motivates a feature-level reinforcement strategy, leading to \textbf{Geometric Residual Gating (GRG)}, which adaptively enhances informative regions while preserving pretrained geometric consistency. Therefore, we propose \textbf{MAETrack}, a lightweight adaptation framework that transfers shallow geometric knowledge while enabling deeper layers to adapt more freely to the downstream tracking objective.

\subsection{Methodology: MAETrack}

To instantiate the proposed structured representation adaptation principle, MAETrack rethinks the conventional fine-tuning paradigm by shifting from full-network parameter inheritance to a Layer-Selective Initialization (LSI) strategy, followed by feature-level modulation. As illustrated in Figure~\ref{fig:maetrack-overview}, MAETrack operates on a single unified tracking backbone rather than a dual-stream architecture, ensuring that representation adaptation is performed within a consistent geometric space. Instead of treating all layers uniformly, the framework explicitly aligns with the layer-wise specialization property of MAE representations by partitioning transferable and task-specific components along network depth. Following this principle, MAETrack is composed of two complementary instantiations: \textbf{LSI}, which preserves transferable shallow geometric priors while decoupling deep reconstruction semantics, and \textbf{Geometric Residual Gating (GRG)}, which performs spatially-aware reinforcement of structurally salient regions in the search BEV features prior to template-search fusion.

\subsection{Layer-Selective Initialization (LSI)}

The core idea behind LSI is motivated by the hierarchical structure of MAE-pretrained representations. Rather than treating all layers as equally transferable, we observe that different network depths encode fundamentally different types of geometric information due to the reconstruction-oriented pre-training objective. From a representation learning perspective, shallow layers tend to encode generalizable low-level geometric primitives, such as occupancy patterns, local surfaces, and boundary structures, which are relatively invariant across tasks and thus more transferable to downstream 3D SOT. In contrast, deeper layers are progressively shaped by the reconstruction objective, learning higher-level semantic abstractions that are optimized for recovering missing content rather than precise instance-level spatial matching.

This layer-wise specialization introduces a structural mismatch when directly transferring all pretrained layers to tracking, since 3D SOT requires accurate geometric correspondence and spatial consistency rather than reconstruction fidelity. Therefore, full-network fine-tuning may lead to interference between transferable geometric representations and reconstruction-specific semantics. To address this, we propose \textbf{Layer-Selective Initialization (LSI)}, which performs \textit{structure-aware transfer along the network depth}. Formally, consider a sparse 3D convolutional backbone partitioned into $L$ stages, denoted by its parameter sequence $\Theta = \{\theta_1, \theta_2, \dots, \theta_L\}$.

Instead of the conventional full-sequence loading strategy, LSI implements a selective inheritance mechanism along the network depth. Specifically, we introduce a binary transfer mask $m_i$ to determine whether the $i$-th backbone stage inherits MAE-pretrained parameters:

\begin{equation}
m_i=
\begin{cases}
1, & i\leq k,\\
0, & i>k,
\end{cases}
\end{equation}

where $k$ denotes the number of inherited shallow stages. The initialized parameters of each stage are then formulated as:

\begin{equation}
\theta_i^{init}
=
m_i\theta_i^{MAE}
+
(1-m_i)\theta_i^{rand},
\end{equation}

where $\theta_i^{MAE}$ represents the parameters from the MAE-pretrained backbone and $\theta_i^{rand}$ denotes random initialization. This formulation explicitly models LSI as a layer-wise parameter transfer operator, where transferable geometric representations are inherited from pre-training while reconstruction-specific deep representations are released for task-specific adaptation.

\noindent\textbf{Shallow Geometric Transfer:} We initialize shallow layers $\Theta_{\text{shallow}} = \{\theta_1, \dots, \theta_k\}$, where $k < L$, using pretrained weights~\cite{lin2024bevmae} to preserve transferable geometric structure. These layers define a low-level geometric subspace that is highly relevant for tracking under sparse observations. In our implementation, the shallow initialization covers the first three backbone stages, and the effect of this design is further validated through stage-wise ablations in the experiments.

\noindent\textbf{Deep Representation Re-initialization:} The deeper layers $\Theta_{\text{deep}} = \{\theta_{k+1}, \dots, \theta_L\}$ are randomly initialized to decouple them from reconstruction-specific biases, so that network rebuild task-specific high-level representations without being constrained by reconstruction-aware semantics. Crucially, after this hybrid initialization, all parameters are jointly optimized by minimizing the tracking objective:

\begin{equation}
\Theta^{*}
=
\arg\min_{\Theta}
\mathcal{L}_{track}(\Theta;\mathbf{P}_{temp},\mathbf{P}_{search}),
\end{equation}

where $\Theta$ includes both inherited shallow layers and randomly initialized deep layers. $\Theta = \Theta_{\text{shallow}} \cup \Theta_{\text{deep}}$ remains fully trainable ($\nabla_{\Theta} \neq 0$) during downstream tracking. Importantly, this design does not freeze any parameters. Instead, after initialization, the entire network remains fully trainable, while the initialization itself constrains the optimization trajectory toward a geometry-preserving subspace. This allows MAETrack to retain transferable shallow geometric priors while avoiding negative transfer from reconstruction-specialized deep representations.

\begin{figure}
\centering
\includegraphics[width=0.9\linewidth]{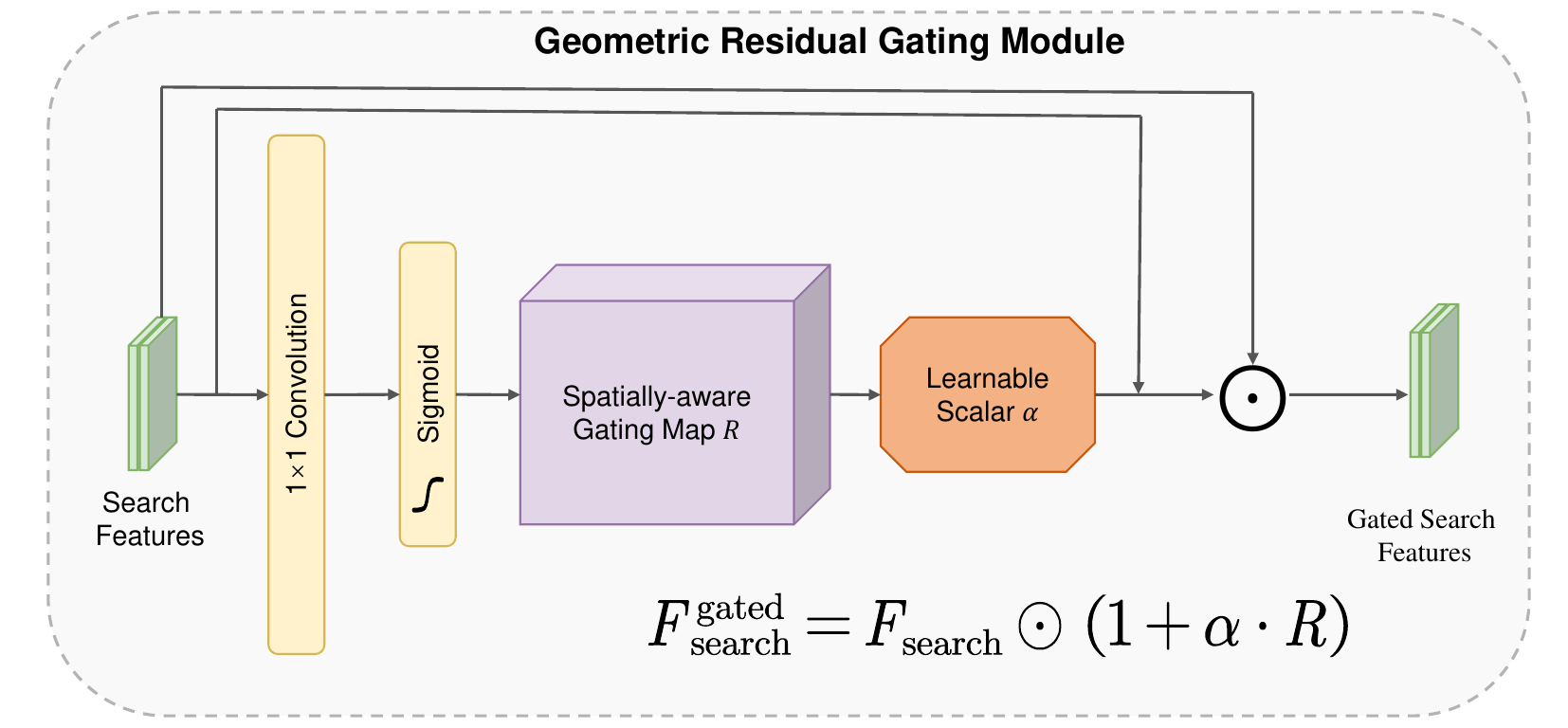}
\caption{Illustration of Geometric Residual Gating (GRG).}
\label{fig:grg-module}
\vspace{-4mm}
\end{figure}

\subsection{Geometric Residual Gating (GRG)}

While Layer-Selective Initialization (LSI) operates at the backbone level to preserve transferable shallow geometric representations, the intermediate feature representations in the downstream tracking pipeline may still suffer from partial degradation due to occlusion, sparsity, and background clutter in the search region. This motivates the need for an explicit feature-level mechanism that reinforces structurally informative regions while preserving the pretrained geometric prior. To address this, we propose \textbf{Geometric Residual Gating (GRG)}, a lightweight geometry-preserving residual modulation module designed specifically for MAE-pretrained feature adaptation in 3D SOT. Unlike conventional spatial attention mechanisms (e.g., SE~\cite{hu2018squeeze} or CBAM~\cite{woo2018cbam}), which perform generic feature reweighting, GRG is designed to selectively enhance structurally salient regions while maintaining the integrity of pretrained geometric representations.

We apply GRG only to the search branch, as it contains spatially ambiguous and noise-corrupted observations due to background clutter and partial occlusion, whereas the template branch provides a relatively stable geometric reference derived from the first-frame annotation. Therefore, feature reinforcement is more critical in the search region, where MAE-pretrained representations are more likely to be degraded by sparse observations.

The key motivation of GRG is that under MAE-pretrained representations and sparse observations, spatial reliability is highly non-uniform. In particular, structurally salient regions such as object boundaries and occupied areas tend to preserve more stable geometric cues, making them more informative for downstream tracking.

Given a pair of point clouds, let 
$F_{\text{search}} \in \mathbb{R}^{C \times H \times W}$
and 
$F_{\text{temp}} \in \mathbb{R}^{C \times H \times W}$
denote the BEV features extracted by the backbone. 
GRG first estimates a spatial geometric reliability map from the search feature:
\begin{equation}
R_s
=
\sigma
(
W_g * F_{\text{search}} + b_g
),
\tag{7}
\end{equation}

where $W_g$ and $b_g$ denote the parameters of a lightweight $1\times1$ convolutional layer and $\sigma(\cdot)$ denotes the Sigmoid activation. Importantly, this gating mechanism does not aim to perform global feature reweighting; instead, it identifies spatial regions where geometric reinforcement is beneficial under sparse and occluded observations. The resulting gate highlights regions where geometric reinforcement is most beneficial for downstream localization. The predicted map $R_s \in \mathbb{R}^{1\times H\times W}$ represents the spatial reliability of geometric structures in the search region.

Instead of directly replacing the original feature representation, GRG introduces a residual geometric enhancement:

\begin{equation}
\Delta F_{\text{search}}
=
\alpha
(
F_{\text{search}}
\odot
R_s
),
\tag{8}
\end{equation}

where $\alpha$ is a learnable scalar controlling the reinforcement strength. 
This residual formulation ensures that the original MAE-pretrained geometric representation remains the dominant component while allowing adaptive enhancement of informative spatial regions.

The final search feature is obtained as:

\begin{equation}
\widetilde F_{\text{search}}
=
F_{\text{search}}
+
\Delta F_{\text{search}}
=
F_{\text{search}}
\odot
(1+\alpha R_s).
\tag{9}
\end{equation}

The adapted search representation is then fused with the template feature:

\begin{equation}
F_{\text{fusion}}
=
\mathcal{F}_{\text{conv\_head}}
(
\operatorname{Concat}
(
F_{\text{template}},
\widetilde F_{\text{search}}
)
).
\end{equation}

From a representation learning perspective, GRG can be interpreted as a geometry-preserving adaptation operator, which enhances task-relevant spatial structures while maintaining consistency with pretrained geometric priors. This is particularly important in 3D SOT, where sparse observations and occlusions can easily weaken local geometric signals. Through the synergy with LSI, GRG enables MAETrack to achieve a unified representation-preserving adaptation framework, where LSI operates at the depth level of the backbone, and GRG operates at the spatial level of feature refinement, jointly ensuring effective transfer of MAE-pretrained geometric knowledge to downstream tracking tasks.

\subsection{Prediction Head and Training Loss.} After template-search fusion, the fused representation is fed into the tracking head to predict the target state. Following the common 3D SOT setting, we directly regress the target center and orientation using the final fused features, while keeping the baseline's part-to-part motion modeling and prediction head strictly unchanged. In other words, our modifications are limited to the backbone initialization strategy and the pre-fusion search-side feature modulation, without altering the downstream motion modeling or supervision design. The entire MAETrack framework is trained end-to-end with the same tracking objective as the baseline. Specifically, we adopt the standard regression losses used in prior 3D SOT works~\cite{v2b}:
\begin{equation}
\mathbf{L}_{\text{track}}=\lambda_1\mathcal{L}_{(x,y)}+\lambda_2\mathcal{L}_{z}+\lambda_3\mathcal{L}_{rot},
\end{equation}
where $\lambda_1$, $\lambda_2$, and $\lambda_3$ balance the position and orientation terms. This design keeps the supervision protocol unchanged and isolates the effect of our transfer strategy to the backbone initialization and search-side feature modulation.

\section{Experiments}\label{sec:exp}
\noindent\textbf{Dataset and Metrics.} 
We follow the common setup~\cite{p2p,hu2025mvctrack} and conduct experiments on the KITTI~\cite{kitti} and large-scale nuScenes~\cite{nuScenes} dataset. Notably, due to the limited data size of the KITTI (only 19 training, 2 validation sequences) makes it challenging to adequately evaluate the methods. In contrast, the nuScenes dataset comprises 700 training and 150 validation sequences, across 40K point cloud frames, allowing for a more comprehensive evaluation. The evaluation metrics is followed the common setup~\cite{ptt,ptt-journal} to report \textit{Success} and \textit{Precision} based on one pass evaluation (OPE)~\cite{otb2013,kristan2016novel}.

\noindent\textbf{Implementation Details.}
Following prior 3D SOT methods~\cite{p2b,ptt-journal}, we use frame \(t-1\) as the template and frame \(t\) as the search frame. The category-dependent crop ranges are $[(-4.8, 4.8), (-4.8, 4.8), (-1.5, 1.5)]$  for cars, $[(-1.92, 1.92), (-1.92, 1.92), (-1.5, 1.5)]$ for pedestrians, and $[(-9.6,9.6), (-9.6,9.6), (-3.0,3.0)]$ for trucks, trailers, and buses. Their voxel sizes are \((0.075,0.075,0.15)\), \((0.03,0.03,0.15)\), and \((0.15,0.15,0.30)\) , respectively. Each frame pair generates four candidates through reference-box perturbation, and synchronized horizontal flipping is additionally used on nuScenes. For LSI, the input convolution and the first two residual sparse-convolution stages are initialized from the Waymo-pretrained BEV-MAE checkpoint, while the deeper stages are randomly initialized; all parameters remain trainable. The GRG coefficient \(\alpha\) is initialized to \(10^{-3}\). We use AdamW with an initial learning rate of \(2\times10^{-4}\), weight decay \(10^{-5}\), cosine annealing to \(10^{-6}\), mixed-precision training, and gradient clipping with a maximum norm of 35. The global batch size is 256 on one RTX 4090 GPU. The nuScenes car/pedestrian, truck/bus, and trailer models are trained for 20, 50, and 100 epochs, respectively, while the KITTI models are trained for 50 epochs. Validation is performed every epoch, and the checkpoint with the best Precision is selected. In our stage notation, Stage 1 denotes the input sparse convolution block, while Stages 2--4 denote the subsequent residual sparse-convolution stages. Thus, Stage 1+2+3 corresponds to initializing the input convolution and the first two residual stages from the BEV-MAE~\cite{lin2024bevmae} pre-trained checkpoint. 

\begin{algorithm}[!htbp]
    \caption{Tracking process of \textbf{MAETrack}}
    \label{alg:maetrack}
    \KwIn{
    A point-cloud sequence $\{\mathcal{P}_t\}_{t=1}^{T}$; 
    an initial target bounding box 
    $\mathbf{B}_1=[x_1,y_1,z_1,h,w,l,\theta_1]^T$ in the first frame; 
    MAE Pre-trained weights $\Theta^{\mathrm{MAE}}$; 
    the number of inherited shallow stages $k$.
    }
    \KwOut{
    A series of predicted bounding boxes 
    $\{\mathbf{B}_t=[x_t,y_t,z_t,h,w,l,\theta_t]^T\}_{t=2}^{T}$.
    }
    \textbf{Offline model construction with LSI:}\\
    Build a tracking backbone $\Phi(\cdot;\Theta)$ with $L$ stages, where 
    $\Theta=\{\theta_1,\theta_2,\ldots,\theta_L\}$;
    Initialize shallow stages $\{\theta_1,\ldots,\theta_k\}$ from 
    $\Theta^{\mathrm{MAE}}$ and randomly initialize deeper stages 
    $\{\theta_{k+1},\ldots,\theta_L\}$; Keep all parameters trainable and optimize the whole \textbf{MAETrack} model with the tracking loss $\mathcal{L}_{\mathrm{track}}$.\;

    \textbf{Online tracking:}\\
    \For{$t=2$ \KwTo $T$}
    {
        Crop the template point cloud $\mathcal{P}_{\mathrm{temp}}^{t-1}$ 
        from frame $t-1$ according to $\mathbf{B}_{t-1}$ and the search point cloud $\mathcal{P}_{\mathrm{search}}^{t}$ 
        from frame $t$ around the previous prediction $\mathbf{B}_{t-1}$. And convert $\mathcal{P}_{\mathrm{temp}}^{t-1}$ and 
        $\mathcal{P}_{\mathrm{search}}^{t}$ into voxelized BEV inputs.\;

        Extract template and search features using the LSI-initialized backbone:
        \vspace{-2mm}
        \[
        \mathbf{F}_{\mathrm{template}},\mathbf{F}_{\mathrm{search}}
        =
        \Phi(\mathcal{P}_{\mathrm{temp}}^{t-1}),
        \Phi(\mathcal{P}_{\mathrm{search}}^{t}).
        \]

        Predict the spatial gating map from the search feature:
        \vspace{-2mm}
        \[
        \mathbf{R}
        =
        \sigma\big(\mathcal{F}_{\mathrm{gate}}(\mathbf{F}_{\mathrm{search}})\big).
        \]

        Apply Geometric Residual Gating (GRG) to obtain enhanced search feature:
        \vspace{-2mm}
        \[
        \mathbf{F}_{\mathrm{search}}^{\mathrm{gated}}
        =
        \mathbf{F}_{\mathrm{search}}
        \odot
        \big(1+\alpha\mathbf{R}\big).
        \]

        Fuse the template feature and the gated search feature:
        \vspace{-2mm}
        \[
        \mathbf{F}_{\mathrm{fusion}}
        =
        \mathcal{F}_{\mathrm{conv\_head}}
        \big(
        \mathrm{Concat}(
        \mathbf{F}_{\mathrm{template}},
        \mathbf{F}_{\mathrm{search}}^{\mathrm{gated}})
        \big).
        \]

        Predict the target center and heading angle using the tracking head and keep the target size $(h,w,l)$ from the template and obtain: $\mathbf{B}_t=[\hat{x}_t,\hat{y}_t,\hat{z}_t,h,w,l,\hat{\theta}_t]^T.$

    }
    \textbf{Return} $\{\mathbf{B}_t\}_{t=2}^{T}$.
\end{algorithm}

\noindent\textbf{Inference.} During inference stage, MAETrack predicts relative motion $[\Delta x,\Delta y,\Delta z,\Delta r]$ frame-by-frame. The regressed relative motion is then applied to target box $\mathcal{B}_{t-1}$ in the previous frame to locate the target box $\mathcal{B}_{t}$ in current frame. Alg.~\ref{alg:maetrack} presents the whole inference process of MAETrack in one trajectory. Our experiment video is available at Appendix.

\begin{table}[!htbp]
\caption{Comparisons with state-of-the-art methods on KITTI dataset~\cite{kitti}. \textit{Success} / \textit{Precision} are used for evaluation. \textbf{Bold} and \underline{underline} denote the best result and the second-best one respectively. $\dagger$ means re-implementation based on official code.}
\centering
    \resizebox{0.99\textwidth}{!}{
    \normalsize
    \begin{tabular}{c|cc|c|cccc}
          \toprule[0.4mm]
             & & &Mean& Car & Pedestrian &  Van & Cyclist \\
           \multirow{-2}{*}{Paradigm}&\multirow{-2}{*}{Tracker} & \multirow{-2}{*}{Source} & (14,068) & (6,424)&(6,088) & (1,248) & (308) \\
          \midrule
          \multirow{17}{*}{Scratch}&SC3D~\cite{sc3d}& CVPR'19 & 31.2 / 48.5 &41.3 / 57.9   & 18.2 / 37.8 & 40.4 / 47.0 & 41.5 / 70.4   \\ 
          &P2B~\cite{p2b}& CVPR'20& 42.4 / 60.0 & 56.2 / 72.8 & 28.7 / 49.6 & 40.8 / 48.4 & 32.1 / 44.7   \\ 
          &3D-SiamRPN\cite{3dsiamrpn} & IEEE Sensors J &46.6 / 64.9  & 58.2 / 76.2 & 35.2 / 56.2 & 45.7 / 52.9 & 36.2 / 49.0 \\ 
        &PTT~\cite{shan2021ptt}&IROS'21 & 55.1 / 74.2 &67.8 / 81.8& 44.9 / 72.0 &43.6 / 52.5& 37.2 / 47.3 \\
        &V2B~\cite{v2b}&NeurIPS'21 & 58.4 / 75.2 &70.5 / 81.3   & 48.3 / 73.5 & 50.1 / 58.0 & 40.8 / 49.7   \\ 
        &PTTR~\cite{pttr}& CVPR'22& 57.9 / 78.2 &65.2 / 77.4   & 50.9 / 81.6 & 52.5 / 61.8 & 65.1 / 90.5  \\ 
        &M$^2$Track~\cite{m2track} & CVPR'22& 62.9 / 83.4 &65.5 / 80.8   & 61.5 / 88.2 & 53.8 / 70.7 & 73.2 / 93.5 \\ 
          &M$^2$Track++~\cite{m2track++} & TPAMI'23& 66.5 / 85.2 &71.1 / 82.7  & 61.8 / 88.7 & 62.8 / 78.5 & 75.9 / 94.0  \\ 
        &GLT-T~\cite{glt}&AAAI'23 & 60.1 / 79.3 &68.2 / 82.1   & 52.4 / 78.8 & 52.6 / 62.9 & 68.9 / 92.1\\ 
        &MBPTrack~\cite{mbptrack} & ICCV'23 & 70.3 / 87.9 & 73.4 / 84.8 &68.6 / 93.9 &61.3 / 72.7& \underline{76.7} / 94.3 \\
        &SyncTrack~\cite{synctrack} &ICCV'23 & 64.1 / 81.9 &73.3 / 85.0 & 54.7 / 80.5&60.3 / 70.0& 73.1 / 93.8 \\
          & PTTR++~\cite{pttr++} & TPAMI'24 & 63.9 / 82.8 &\underline{73.4} / 84.5 &55.2 / 84.7 &55.1 / 62.2& 71.6 / 92.8 \\ 
         &VoxelTrack~\cite{lu2024voxeltrack} & ACM MM'24&70.4 / 88.3 &72.5 / 84.7 & 67.8 / 92.6  & 69.8 / \underline{83.6} & 75.1 / \textit{94.7}\\
         &VPMCAN~\cite{zhao2025vpmcan} &PR'25 &70.1 / 88.6 &74.6 / \underline{86.3} &65.8 / \textbf{96.2} &63.1 / 73.5 &\textbf{76.9} / \textbf{96.6} \\

         \hline
         &P2P$^\dagger$~\cite{p2p} & IJCV'25& \underline{71.1} / \underline{88.9} 
         & 73.2 / 85.4
         & \textbf{69.0} / \underline{93.4} 
         & \underline{69.8} / 83.3 
         & 75.1 / 94.0 \\ 
        &\textbf{MAETrack} & \textbf{Ours}&\textbf{72.0} / \textbf{89.9} & \textbf{75.2} / \textbf{87.5}&\underline{68.8} / 93.2 &\textbf{70.5} / \textbf{84.6} &75.5 / \underline{94.7} \\ 
          \bottomrule[0.4mm]
    \end{tabular}
    }
\label{tab:KITTI}
\vspace{-4mm}
\end{table}

\begin{table}[!htbp]
\centering
\caption{Comparison with SOTA methods on the nuScenes dataset. Success and Precision are used for evaluation, \textbf{Bold} and \underline{underline} denote the best result and the second-best one respectively. * 64159 indicates the number of instances of cars.}
\resizebox{0.99\linewidth}{!}{
\begin{tabular}{c|cccccc}
\toprule
\multirow{2}{*}{Method}& Mean &Car& Pedestrian& Truck& Trailer& Bus\\
&(117,278) &(64,159)*&(33,227)&(13,587)&(3,352)&(2,953)\\
\midrule
SC3D~\cite{sc3d}& 20.70 / 20.20&22.31 / 21.93 & 11.29 / 12.65 & 35.28 / 28.12 & 35.28 / 28.12 & 29.35 / 24.08 \\
P2B~\cite{p2b} & 36.48 / 45.08 &38.81 / 43.81 & 28.39 / 52.24 & 48.96 / 40.05 & 48.96 / 40.05 & 32.95 / 27.41 \\
PTT~\cite{ptt-journal}& 36.33 / 41.72 &41.22 / 45.26 & 19.33 / 32.03 & 50.23 / 48.56 & 51.70 / 46.50 & 39.40 / 36.70 \\
BAT~\cite{bat} & 38.10 / 45.71 &40.73 / 43.29 & 28.83 / 53.32 & 52.59 / 44.89 & 52.59 / 44.89 & 35.44 / 28.01 \\
V2B~\cite{v2b} & - / - & 54.40 / 59.70 & 30.10 / 55.40 & 53.70 / 54.50 & 54.90 / 51.44 & - / - \\
M$^2$-Track~\cite{m2track}& 49.23 / 62.73 &55.85 / 65.09 & 32.10 / 60.92 & 57.36 / 59.54 & 57.61 / 58.26 & 51.39 / 51.44 \\
PTTR~\cite{pttr}& 44.50 / 52.07 &51.89 / 58.61 & 29.90 / 45.09 & 45.30 / 44.74 & 45.87 / 38.36 & 43.14 / 37.74 \\
GLT-T~\cite{glt}& 44.42 / 54.33 &48.52 / 54.29 & 31.74 / 56.49 & 52.74 / 51.43 & 57.60 / 52.01 & 44.55 / 40.69 \\
PTTR++~\cite{pttr++}& 51.86 / 60.63 &59.96 / 66.73& 32.49 / 50.50 & 59.85 / 61.20& 54.51 / 50.28& 53.98 / 51.22\\
VoxelTrack~\cite{lu2024voxeltrack} & 59.00 / \underline{71.40} & 63.90 / 71.60 & \underline{46.80} / \underline{75.90} & \underline{64.80} / \underline{65.90} & 69.50 / 64.30 & \underline{60.10} / \underline{57.70} \\
\midrule
 P2P$^\dagger$~\cite{p2p}  &\underline{59.22} / 71.19 &\underline{64.61} / \underline{71.98} 
           & 45.64 / 74.62
           & 64.42 / 65.37 
           & \underline{70.23} / \underline{66.08} 
           & 58.54 / 56.13
           \\
            \textbf{MAETrack} &\textbf{61.16}  /  \textbf{73.54}&
            \textbf{66.05}  /  \textbf{73.57}& \textbf{47.21}  /  \textbf{75.96}& \textbf{68.50}  /  \textbf{70.53}& \textbf{74.10} /  \textbf{71.63}& \textbf{62.11}  /  \textbf{60.86}  \\

\bottomrule
\end{tabular}
}
\label{tab:NuScenes}
\end{table}

\subsection{Comparison with SOTA Method}

\noindent\textbf{Results on KITTI Dataset.} As summarized in Table~\ref{tab:KITTI}, we evaluate our MAETrack on the KITTI dataset and compare it with state-of-the-art 3D trackers. It is worth noting that KITTI poses a unique challenge due to its extremely limited training data size (only 19 sequences), which typically makes data-hungry models prone to severe overfitting. Despite this, our MAETrack achieves compelling performance. Specifically, on the Car category, MAETrack attains \textbf{75.2\%}/\textbf{87.5\%} in Success and Precision, respectively, outperforming the strong baseline tracker P2P by \textbf{+2.0\%}/\textbf{+2.1\%}. This significant performance gain explicitly demonstrates the strong data efficiency of our framework: the geometric prior knowledge successfully transferred by our MAE pre-training effectively compensates for the lack of large-scale annotated data, providing a structurally robust representation that prevents the model from overfitting to sparse training samples.

\noindent\textbf{Results on nuScenes Dataset.} Table~\ref{tab:NuScenes} summarizes the tracking results of our MAETrack and other state-of-the-art methods on the large-scale nuScenes dataset~\cite{nuScenes}. Compared to our strong baseline P2P~\cite{p2p}, our MAETrack achieves comprehensive improvements across almost all categories. Specifically, on the Car category, our method reaches 66.05\%/73.57\%, and on the challenging rigid classes like Truck and Trailer, we achieve significant boosts of +4.38\%/+5.16\% and +3.87\%/+5.55\%, respectively. Notably, on the Bus category, our method yields a large gain, reaching an improvement of +3.57\% / +4.73\% over the baseline. Overall, MAETrack surpasses previous methods and demonstrates that our layer-selective initialization and geometric residual gating effectively transfer geometric priors to significantly enhance tracking performance.

\begin{table}[!htbp]
\caption{Comparison of the running speeds on some representative methods.}
\label{table:speed}
\resizebox{\linewidth}{!}{
\begin{tabular}{c|cccccc}
\toprule[.05cm]
Method & P2B\cite{p2b} & PTT\cite{ptt-journal} & BAT\cite{bat}&PTTR~\cite{pttr} & M$^2$Track\cite{m2track} &GLT-T~\cite{glt}\\ 
FPS    & 40.0 & 40.0 & 57.0 &43.0 & 51.2 & 30.0\\ 
\midrule
Method & VoxelTrack\cite{lu2024voxeltrack} & M$^2$Track++\cite{m2track++} & BAT~\cite{bat} &V2B~\cite{v2b} &P2P~\cite{p2p} & \textbf{MAETrack} (our)\\ 
 FPS   & 36.0 & 57.0 & 57.0& 37.0&\textbf{88.9}&  84.3 \\ 
\bottomrule[.05cm]
\end{tabular}
}
\vspace{-1mm}
\end{table}

\noindent\textbf{Running Speed:}
Inference speed is also a vital factor for practical applications. We present a comprehensive speed comparison of MAETrack with other methods in Tab.~\ref{table:speed}. Following common evaluation protocols~\cite{p2b, ptt-journal,p2p}, speed is measured by calculating the average running time of all frames in the Car category. On a single NVIDIA RTX 4090 GPU, MAETrack achieves 84 FPS. Despite the computational overhead incurred by our GRG module, MAETrack yields significant performance improvements, maintaining a better trade-off between accuracy and speed.

\begin{table}[!htbp]
\vspace{-2mm}
\centering
\caption{Comparison on varying levels of sparsity on Car category of KITTI~\citep{kitti} and NuScenes~\citep{nuScenes}. \textbf{Bold} denotes the best result.}
    \resizebox{0.99\linewidth}{!}{
    \normalsize
    \begin{tabular}{c|cccccc}
          \toprule[0.4mm]
        Dataset & \multicolumn{6}{c}{KITTI Car}  \\
          \midrule
        Interval & [0, 10) & [10, 20) & [20, 30) & [30, 40) & [40, 50) & [50, +$\infty$) \\
          \midrule
        Sequence Number & 46  & 29  & 19 & 5  & 3 & 18 \\
          \midrule
        Frame Number & 2,394 & 1,590 &  709 &  97 & 80 & 1,554 \\
          \midrule
          M$^2$Track~\citep{m2track}&  53.0 / 67.1 & 60.2 / 73.9& 62.6 / 75.7 & 77.6 / 92.1 &61.6 / 72.1 & 79.6 / 92.1\\
         P2P~\cite{p2p}&64.8 / 75.0&70.7 / 83.6&73.5 / 87.8 &77.0 / 90.4&\textbf{80.2} / \textbf{92.7} & 81.2 / 92.7 \\
           MAETrack &\textbf{67.1} / \textbf{77.7}&\textbf{71.7} / \textbf{85.8}&\textbf{76.4} / \textbf{89.4}& \textbf{83.3} / \textbf{94.1}& 64.0 / 71.9 & \textbf{81.5} / \textbf{93.3} \\
           \midrule
           \midrule
            Dataset & \multicolumn{6}{c}{NuScenes Car} \\
          \midrule
          Interval &  [0, 10) & [10, 20) & [20, 30) & [30, 40) & [40, 50) & [50, +$\infty$)\\
          \midrule
         Sequence Number& 2,884 & 212  & 101  & 64  & 37  & 362\\
          \midrule
          Frame Number &  45,322& 4,375 &  2,190 & 1,321 & 814 &  10,127\\
          \midrule
          M$^2$Track~\citep{m2track} & 52.1 / 60.9&57.1 / 65.9 & 65.8 / \textbf{73.4} &  68.1 / 76.3& 70.1 / 78.6 & \textbf{75.7} / \textbf{83.1} \\
           P2P~\cite{p2p} & 63.3 / 71.4 & 64.1 / 71.9 & 64.3 / 70.2&69.6 / 77.4 & \textbf{73.9} / 79.9& 72.8 / 79.4\\
           MAETrack& \textbf{64.3} / \textbf{71.7}&\textbf{65.4} / \textbf{73.5} & \textbf{65.9} / 72.7&\textbf{69.7} / \textbf{77.5} & 73.7 / \textbf{81.7}& 72.9 / 80.6\\
          \bottomrule[0.4mm] 
    \end{tabular}}
\label{table10}
\vspace{-5mm}
\end{table}

\noindent\textbf{Robustness to Point-cloud Sparsity.}
To directly validate the robustness of MAETrack under sparse or occluded scenes, we conduct a sparsity analysis on the Car category of KITTI and nuScenes. Specifically, we group tracking sequences according to the number of target points in the first frame and evaluate all frames within each group. The number of sequences and frames in each sparsity interval is also reported in Table~\ref{table10} to reflect the statistical reliability of each group. As shown in Table~\ref{table10}, MAETrack consistently improves over the strong baseline P2P in most sparse intervals on both datasets. On KITTI, MAETrack achieves clear gains in the highly sparse intervals, improving P2P from 64.8/75.0 to 67.1/77.7 in the [0,10) interval and from 70.7/83.6 to 71.7/85.8 in the [10,20) interval. Similar improvements are observed in the [20,30) and [30,40) intervals, where MAETrack obtains 76.4/89.4 and 83.3/94.1, respectively. On nuScenes, where sparse observations are much more common, MAETrack also shows consistent advantages in low-point regimes, improving P2P from 63.3/71.4 to 64.3/71.7 in the [0,10) interval and from 64.1/71.9 to 65.4/73.5 in the [10,20) interval. These results demonstrate that the proposed LSI and GRG modules are particularly beneficial when the available geometric evidence is weak.

It is also worth noting that the performance gap becomes smaller in dense intervals, where baseline trackers already receive sufficient geometric information from the point cloud. This observation is consistent with our motivation: MAE-pretrained shallow geometric priors and search-side residual gating are most useful when the target observation is sparse, partially missing, or structurally ambiguous. The only exception appears in the [40,50) interval on KITTI, where MAETrack underperforms P2P. However, this interval contains only 3 sequences and 80 frames, making the result statistically less stable. Overall, the sparsity analysis provides direct evidence that MAETrack improves robustness under sparse point-cloud observations rather than only improving category-level average performance.


\subsection{Ablation Study}
To validate the effectiveness of our MAETrack, we conduct comprehensive ablation studies on KITTI and nuScenes dataset of Car and Pedestrian category. We adopt the strong P2P as baseline and evaluate the contributions of our core designs: Layer-Selective Initialization (LSI) and Geometric Residual Gating (GRG).

\begin{table}[!htbp]
\caption{Ablation study of the proposed module on KITTI and nuScenes dataset.}
\centering
\resizebox{0.9\linewidth}{!}{
\normalsize
\begin{tabular}{c|cc|cc|cc}
\toprule[0.4mm]
\multirow{2}{*}{Methods} & \multicolumn{2}{c|}{Module} & \multicolumn{2}{c}{KITTI} & \multicolumn{2}{c}{nuScenes}\\
\cmidrule{2-7}
 & LSI & GRG & Car & Ped & Car & Ped\\
\midrule
Baseline & \ding{55} & \ding{55} & 73.20 / 85.40 & \textbf{69.00} / \textbf{93.40} & 64.61 / 71.98 & 45.64 / 74.62\\ \hline
Variant A & \checkmark &\ding{55} & 74.10 / 86.20 & 68.70 / 93.10 & 65.22 / 72.43 & 46.27 / 75.10\\
Variant B & \ding{55} & \checkmark & 73.90 / 86.40 & 68.50 / 93.20 & 65.09 / 72.78 & 46.10 / 75.38\\
MAETrack & \checkmark & \checkmark & \textbf{75.20} / \textbf{87.50} & 68.80 / 93.20  & \textbf{66.05} / \textbf{73.57} & \textbf{47.21} / \textbf{75.96}\\
\bottomrule[0.4mm]
\end{tabular}
}
\label{tab:ablation_overall}
\vspace{-2mm}
\end{table}

\subsubsection{Overall Module Ablation.}
We first investigate the contribution of the proposed LSI and GRG modules on both KITTI and nuScenes. As shown in Table~\ref{tab:ablation_overall}, the proposed modules bring consistent improvements on the large-scale nuScenes benchmark. Specifically, on the Car category, LSI alone improves the baseline from 64.61/71.98 to 65.22/72.43, while GRG alone improves it to 65.09/72.78. Combining both modules further boosts the performance to 66.05/73.57. Similar trends can also be observed on the Pedestrian category, where MAETrack improves the baseline from 45.64/74.62 to 47.21/75.96. These results demonstrate that LSI and GRG are complementary: LSI reduces negative transfer from reconstruction-specific deep representations, while GRG reinforces structurally informative regions in the search feature.

On KITTI, MAETrack also significantly improves the Car category from 73.2/85.4 to 75.2/87.5, showing that the proposed transfer strategy is effective under limited-data settings. However, the gain on KITTI Pedestrian is less obvious, where the baseline remains slightly better. We attribute this to the small scale and category bias of KITTI, where the pedestrian subset contains highly sparse and non-rigid targets with limited training diversity. Overall, the ablation results indicate that LSI and GRG jointly provide stable improvements, especially on the larger and more diverse nuScenes benchmark.

\begin{table}[!htbp]
\caption{Impact of different initialization strategies. Here, ``Only Stage $i$'' means that only the $i$-th backbone stage is initialized from the BEV-MAE checkpoint, while all other stages are randomly initialized. ``Stage 1+2+3'' denotes cumulative initialization of the first three stages.}
\centering
\resizebox{0.9\linewidth}{!}{
\normalsize
\begin{tabular}{l|cc|cc}
\toprule[0.4mm]
\multirow{2}{*}{Initialization Strategy} & \multicolumn{2}{c}{KITTI} & \multicolumn{2}{c}{nuScenes}\\
\cmidrule{2-5}
& Car & Ped & Car & Ped \\
\midrule
Baseline & 73.20 / 85.40 & \textbf{69.00} / \textbf{93.40} & 64.61 / 71.98 & 45.64 / 74.62\\
Full Network Init. & 73.80 / 86.10 & 68.30 / 92.80 & 65.09 / 72.58 & 46.22 / 75.24\\
\midrule
Only Stage 1 & 74.40 / 86.50 & 68.60 / 93.00 & 65.39 / 72.90 & 46.13 / 75.11\\
Only Stage 2 & 73.60 / 85.90 & 68.40 / 92.90 & 64.69 / 71.56 & 46.42 / 75.23\\
Only Stage 3 & 72.90 / 85.10 & 68.20 / 92.70 & 63.65 / 71.23 & 46.22 / 74.45\\
Only Stage 4 & 72.40 / 84.60 & 68.00 / 92.60 & 63.14 / 69.60 & 46.01 / 75.00\\
\midrule
Stage 1+2 & 75.00 / 87.20 & 68.70 / 93.10 & \textbf{66.10} / 73.01 & 46.58 / 75.42\\
Stage 1+2+3 (LSI) & \textbf{75.20} / \textbf{87.50} & 68.80 / 93.20 & 66.05 / \textbf{73.57} & \textbf{47.21} / \textbf{75.96}\\
\bottomrule[0.4mm]
\end{tabular}
}
\label{tab:ablation_lsi}
\vspace{-2mm}
\end{table}

\subsubsection{Effect of Layer-Selective Initialization.}
A core premise of MAETrack is the layer-wise transfer mismatch between reconstruction-based pre-training and downstream tracking. To examine this hypothesis more directly, we compare different initialization strategies in Table~\ref{tab:ablation_lsi}, including training from scratch, full-network initialization, single-stage initialization, and cumulative stage initialization. Here, ``Only Stage $i$'' indicates that only the $i$-th backbone stage is initialized from the BEV-MAE checkpoint, while all other stages are randomly initialized. The results reveal clear layer-wise transfer differences. First, full-network initialization brings only limited gains over the baseline. For example, on nuScenes Car, it improves the baseline from 64.61/71.98 to 65.09/72.58, while on KITTI Car it improves from 73.20/85.40 to 73.80/86.10. This indicates that simply loading all pre-trained weights is not sufficient for effective tracking adaptation. Second, the single-stage initialization results show that different stages exhibit different transfer behaviors. Initializing early stages generally provides more stable improvements, while initializing deeper stages alone yields limited or even degraded performance, especially on the Car category. For instance, on nuScenes Car, Only Stage 1 achieves 65.39/72.90, whereas Only Stage 3 and Only Stage 4 drop to 63.65/71.23 and 63.14/69.60, respectively. This observation supports our hypothesis that shallow layers preserve more transferable geometric cues, whereas deeper layers are more specialized to the reconstruction pretext task.

Furthermore, cumulative shallow-stage initialization provides a better overall transfer strategy. On KITTI Car, Stage 1+2+3 improves the baseline from 73.20/85.40 to 75.20/87.50. On nuScenes, Stage 1+2 obtains the highest Car Success, while Stage 1+2+3 achieves the best Car Precision and the best Pedestrian performance. Compared with full-network initialization, the default LSI setting improves nuScenes Car from 65.09/72.58 to 66.05/73.57 and Pedestrian from 46.22/75.24 to 47.21/75.96. These results indicate that the benefit of MAE pre-training does not simply come from using pre-trained weights, but from selectively inheriting layers that are more suitable for 3D SOT. Therefore, LSI provides a more appropriate adaptation strategy by preserving transferable shallow geometric priors while avoiding excessive dependence on reconstruction-specialized deep representations. It is worth noting that on KITTI Pedestrian, the baseline remains slightly better than the initialization-based variants. This suggests that the benefit of MAE transfer may vary across categories and datasets, especially under limited data and highly non-rigid pedestrian motion. Nevertheless, the consistent gains on KITTI Car and the larger-scale nuScenes benchmark demonstrate the effectiveness of layer-selective transfer for adapting MAE-pretrained representations to 3D tracking.

\begin{table}[!htbp]
\centering
\caption{Layer-wise representation diagnosis of BEV-MAE pre-trained features on nuScenes Car and Ped. 
CKA measures the similarity between BEV-MAE pre-trained features and tracking fine-tuned features at each stage. 
Foreground F1 and Boundary IoU are obtained by training a lightweight linear probe on frozen stage features to predict target occupancy and boundary regions in BEV space.}
\label{tab:cka_probe}
\setlength{\tabcolsep}{4.5pt}
\renewcommand{\arraystretch}{1.05}
\resizebox{0.9\linewidth}{!}{
\begin{tabular}{c|c|cc|c|c}
\toprule[0.4mm]
Stage & CKA Similarity $\uparrow$ & Foreground F1 $\uparrow$ & Boundary IoU $\uparrow$ & \multicolumn{2}{c}{Only-stage Init.} \\
\midrule
Stage 1 & \textbf{0.84} & \textbf{72.8} & \textbf{44.6} & \textbf{65.39} / \textbf{72.90} & 46.13 / 75.11\\
Stage 2 & 0.71 & 70.9 & 42.7 & 64.69 / 71.56 & \textbf{46.42} / \textbf{75.23} \\
Stage 3 & 0.52 & 66.3 & 38.9 & 63.65 / 71.23 & 46.22 / 74.45 \\
Stage 4 & 0.36 & 62.5 & 35.1 & 63.14 / 69.60 & 46.01 / 75.00\\
\bottomrule[0.4mm]
\end{tabular}
}
\vspace{-3mm}
\end{table}

\subsubsection{Layer-wise Representation Diagnosis.}
Although the initialization ablation in Table~\ref{tab:ablation_lsi} provides performance-level evidence for the layer-wise transfer mismatch, it does not directly reveal what type of information is preserved at different stages. To further analyze the representation behavior of BEV-MAE pre-trained features, we conduct a layer-wise representation diagnosis on the nuScenes Car category.

Specifically, we evaluate two complementary aspects. First, we compute the linear CKA similarity between BEV-MAE pre-trained features and tracking fine-tuned features at each stage. Given two feature matrices $X$ and $Y$ extracted from the same stage of the pre-trained and fine-tuned backbones, respectively, the CKA similarity is computed as:
\begin{equation}
\mathrm{CKA}(X,Y)=
\frac{\|X^{\top}Y\|_F^2}
{\|X^{\top}X\|_F \|Y^{\top}Y\|_F}.
\end{equation}
A higher CKA value indicates that the representation is more consistently preserved after tracking fine-tuning. Second, we perform a lightweight geometric probing experiment. For each frozen stage feature, we train a $1\times1$ linear probe to predict two BEV-level geometric targets derived from the ground-truth 3D bounding box: a foreground occupancy mask and a boundary mask. Foreground F1 and Boundary IoU are used to evaluate whether the corresponding stage preserves tracking-relevant local geometric information. As shown in Table~\ref{tab:cka_probe}, shallow stages exhibit substantially higher CKA similarity than deeper stages. Stage 1 obtains the highest CKA similarity of 0.84, while Stage 4 decreases to 0.36. This indicates that shallow BEV-MAE representations remain more consistent with the downstream tracking representation, whereas deeper representations undergo larger shifts during fine-tuning. Such behavior suggests that deeper layers are more specialized to the reconstruction pretext task and require stronger adaptation to tracking-specific objectives.

The geometric probing results show a consistent trend. Stage 1 achieves the best Foreground F1 and Boundary IoU, indicating that shallow layers preserve more target-relevant occupancy and boundary structures. In contrast, deeper stages show weaker probing performance, suggesting that they contain less directly transferable local geometric information for precise target localization. These representation-level observations are consistent with the only-stage initialization results in Table~\ref{tab:ablation_lsi}, where Stage 1 provides more stable transfer gains while Stage 3 and Stage 4 alone lead to degraded performance. Together, these results provide stronger evidence for our layer-wise transfer hypothesis: shallow MAE-pretrained layers mainly encode transferable geometric priors, while deeper layers are more reconstruction-specific. This further supports the design of LSI, which selectively inherits shallow stages while allowing deeper layers to adapt to the downstream 3D tracking objective.

\begin{table}[!htbp]
\centering
\caption{Comparison between GRG and generic attention/gating modules on nuScenes dataset. All modules are inserted into the search branch before template-search fusion under the same LSI initialization. $A_c$ and $A_s$ denote channel and spatial attention maps, respectively. $R$ denotes the spatial gating map predicted by the gating module.}
\label{tab:grg_attention_compare}
\setlength{\tabcolsep}{5pt}
\renewcommand{\arraystretch}{1.05}
\resizebox{\linewidth}{!}{
\begin{tabular}{l|c|c|cc|cc}
\toprule[0.4mm]
\multirow{2}{*}{Method} & \multirow{2}{*}{Applied Branch} & \multirow{2}{*}{Modulation Form} & \multicolumn{2}{c|}{Car} & \multicolumn{2}{c}{Pedestrian} \\
 &  &  & Success & Precision & Success & Precision \\
 \midrule
 LSI only & -- & $F_s$ &65.22 & 72.43 & 46.27 & 75.10\\
\midrule
LSI + SE~\cite{hu2018squeeze} 
& Search & $F_s \odot A_c$ 
& 65.36 & 72.66 & 46.38 & 75.22 \\

LSI + CBAM~\cite{woo2018cbam} 
& Search & $F_s \odot A_c \odot A_s$ 
& 65.52 & 72.88 & 46.55 & 75.35 \\

LSI + Spatial-only CBAM~\cite{woo2018cbam} 
& Search & $F_s \odot R$ 
& 65.68 & 73.05 & 46.73 & 75.47 \\

LSI + Residual Spatial Gate~\cite{wang2017residual} 
& Search & $F_s \odot (1 + R)$ 
& 65.86 & 73.31 & 46.98 & 75.71 \\
LSI + GRG (Ours) 
& Search & $F_s \odot (1 + \alpha R)$ 
& \textbf{66.05}  &  \textbf{73.57} & \textbf{47.21}  &  \textbf{75.96} \\
\bottomrule[0.4mm]
\end{tabular}
}
\vspace{-3mm}
\end{table}

\subsubsection{Comparison with Generic Attention and Gating Modules.}
To clarify whether the improvement of GRG simply comes from adding a generic attention module, we compare GRG with representative attention and gating mechanisms, including SE, CBAM, spatial-only CBAM, and residual spatial gating. For a fair comparison, all modules are inserted into the same location, i.e., the search branch before template-search fusion, and are trained under the same LSI initialization strategy. As shown in Table~\ref{tab:grg_attention_compare}, generic attention modules bring only limited improvements over LSI alone. SE mainly recalibrates channel responses and provides marginal gains, while CBAM and spatial-only CBAM further improve performance by introducing spatial modulation. However, directly applying spatial gating is still inferior to residual modulation, indicating that preserving the original MAE-pretrained geometric representation is important. The residual spatial gate improves stability, while the proposed GRG achieves the best performance by introducing a learnable residual strength $\alpha$. These results suggest that GRG is not merely a generic attention block, but a geometry-preserving modulation mechanism tailored for MAE-pretrained 3D tracking features.

\begin{table}[!htbp]
\caption{Ablation on the placement of GRG on KITTI and nuScenes dataset.}
\centering
\resizebox{0.9\linewidth}{!}{
\begin{tabular}{l|cc|cc}
\toprule[0.4mm]
\multirow{2}{*}{GRG Placement } & \multicolumn{2}{c}{KITTI} & \multicolumn{2}{c}{nuScenes}\\
\cmidrule{2-5}
& Car & Ped & Car & Ped\\
\midrule
Template Only & 73.30 / 85.70 & 68.10 / 92.50  & 64.21 / 71.76 & 45.80 / 74.68\\
Both Branches & 74.10 / 86.40 & 68.50 / 92.80  & 64.84 / 72.92 & 46.15 / 75.06\\
After Fusion & 74.60 / 86.90 & 68.60 / 93.00 & 65.52 / 72.81 & 46.56 / 75.37\\
Search Only (Ours) & \textbf{75.20} / \textbf{87.50} & \textbf{68.80} / \textbf{93.20} &\textbf{66.05} / \textbf{73.57} & \textbf{47.21} / \textbf{75.96}\\
\bottomrule[0.4mm]
\end{tabular}
}
\label{tab:ablation_grg}
\vspace{-3mm}
\end{table}

\subsubsection{Design Choices in GRG.}
GRG is designed to reinforce search-side geometric features before template-search fusion. To verify this design, we ablate the placement of GRG in Table~\ref{tab:ablation_grg}, including applying it to the template branch only, both branches, after fusion, and the search branch only. The results show that applying GRG only to the search branch consistently achieves the best performance across both KITTI and nuScenes. For example, on KITTI Car, search-only GRG obtains 75.20/87.50, outperforming template-only, both-branch, and after-fusion variants. Similarly, on nuScenes Car, search-only GRG achieves 66.05/73.57, while applying GRG after fusion only obtains 65.52/72.81. This observation supports our design motivation. The template branch usually provides a relatively stable target reference, since it is cropped around the annotated or previously localized target. Applying gating to the template branch may disturb this reliable geometric reference. This explains why template-only modulation yields inferior results. Applying GRG to both branches partially improves over template-only modulation, but still risks perturbing the template representation.

In contrast, the search branch contains more background clutter, spatial ambiguity, and incomplete observations. Therefore, it benefits more from geometry-preserving residual modulation. Applying GRG after fusion is also sub-optimal, because the search features have already been fused with the template before structural reinforcement is performed. As a result, the fusion module has to match unenhanced search features, making it harder to suppress background noise and emphasize target-related geometry. These results indicate that the effectiveness of GRG does not merely come from adding an extra gating operation. Instead, its benefit depends on applying residual geometric reinforcement to the search representation at the appropriate stage, before template-search fusion. This further supports our view that GRG is a search-oriented geometric modulation mechanism tailored for 3D SOT, rather than a generic spatial attention block.

\begin{figure}[!htbp]
    \centering
    \includegraphics[width=0.8\linewidth]{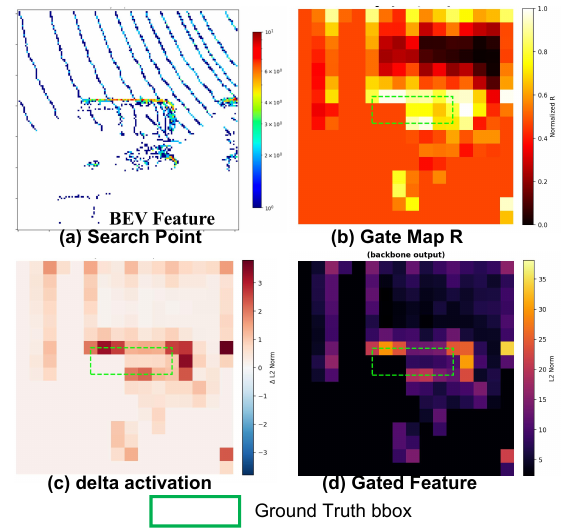}
    \caption{Visualization of GRG feature.}
    \label{fig:visualize_feats}
    \vspace{-3mm}
\end{figure}

\subsection{Visualization results}

\begin{figure}[h]
    \centering
    \includegraphics[width=\linewidth]{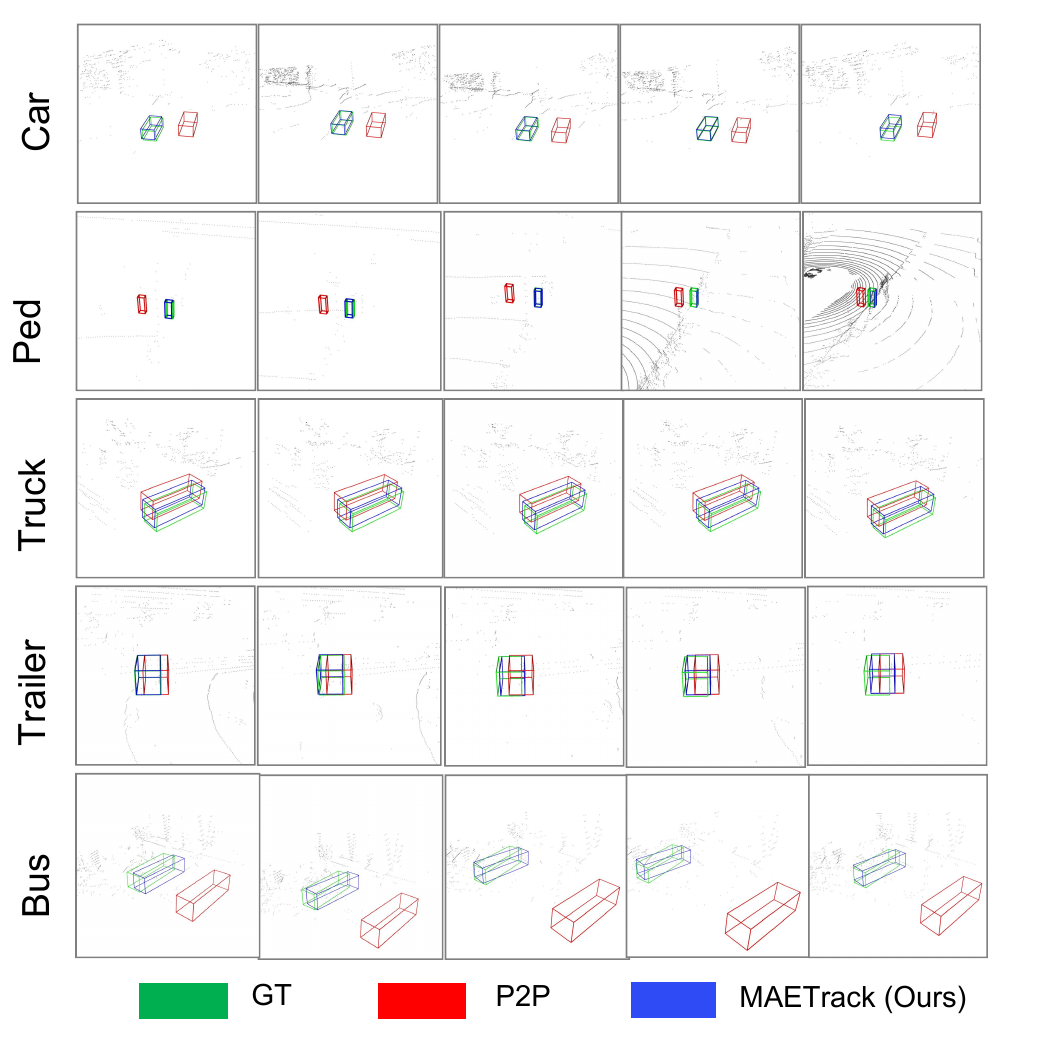}
    \caption{Tracking performance across varying classes on nuScenes~\cite{nuScenes} dataset.}
    \label{fig:visualize}
\end{figure}

\subsubsection{Feature Visualization of GRG} As shown in Figure~\ref{fig:visualize_feats}, visualized results of the Gate Map ($R$) and the corresponding feature variation ($\Delta$ Activation) demonstrate the effectiveness of the GRG module. The modulated weight $R$ ranges from $0.283$ to $0.995$, with peak values densely concentrated along the boundaries of the GT bbox. Consequently, the most significant feature amplifications ($\Delta$ Activation) are precisely aligned with the target's vicinity. This indicates that GRG successfully achieves target-aware spatial modulation, explicitly reinforcing the boundary and local geometric structures of the object while mitigating uniform background distraction.

\subsubsection{Tracking Visualization Across Diverse Object} Fig.~\ref{fig:visualize} provides some visual comparison results over the SOTA baseline P2P~\cite{p2p} on nuScenes~\cite{nuScenes} dataset, across diverse trajectories. Whether in dense scenarios, or sparse scenarios, our MAETrack is able to track the target, while P2P performs inaccurate box estimation. This highlights the superior tracking performance of our MAETrack across a range of conditions, from dense to sparse scenes. We also provide more visualization results in the appendix.

\vspace{-2mm}
\section{Conclusion}\label{sec:conclusion}
\vspace{-2mm}
This paper presents \textbf{MAETrack}, a lightweight framework designed to bridge the gap between reconstructive 3D pre-training and downstream 3D SOT. We identify a layer-wise objective mismatch as a key obstacle:while shallow layers harbor transferable geometric cues, deeper layers specialize in reconstruction, which can impede tracking. To address this, MAETrack employs Layer-Selective Initialization (LSI) to inherit only the foundational geometric weights and Geometric Residual Gating (GRG) to reinforce structurally salient regions via spatial modulation. Benchmarks on KITTI and nuScenes demonstrate that MAETrack significantly outperforms vanilla fine-tuning, particularly for geometrically challenging categories, with negligible overhead. Our findings suggest that the true potential of 3D foundation models lies in their hierarchical geometric priors rather than final outputs. By reconciling these priors with temporal matching, we offer a principled approach to developing efficient, robust robotic perception systems.

\noindent\textbf{Limitations and Future Work.}
Despite its effectiveness, MAETrack has several limitations: \textbf{(1)} MAETrack relies on a single pre-trained source; exploring multi-modal (camera-LiDAR)~\cite{hu2025mvctrack} or heterogeneous pre-training may yield richer priors. \textbf{(2)} GRG currently lacks explicit temporal modeling~\cite{fan2025beyond}, which could be enhanced to handle fast motion or severe occlusion. \textbf{(3)} Our study focuses on CNN-based backbones; extending this paradigm to transformer-based~\cite{2025comptrack,focustrack}. Future work will explore above issues and extending to multi-object tracking to further validate its generalization capabilities.

\vspace{-2mm}
\section*{CrediT authorship contribution statement}
\textbf{Sifan Zhou:} Investigation, Methodology, Software, Validation, Writing - original draft, Writing - review \& editing, Project administration. \textbf{Qiwei Wang:} Methodology, Validation, Visualization, Writing - original draft, Writing - review \& editing. \textbf{Linyue Tan:} Software, Formal analysis, Validation, Writing -original draft \& editing. \textbf{Ziyu Liu:} Writing - original draft \& editing. \textbf{Ziyu Zhao:} Writing - review \& editing. \textbf{Ziyu Zhao:} Formal analysis, Writing - review \& editing. \textbf{Xiaobo Lu:} Methodology, Writing - review, Project administration, Funding acquisition.

\vspace{-2mm}
\section*{Acknowledgment}
This work was supported by the National Natural Science Foundation of China (No. 62271143), the Frontier Technologies R\&D Program of Jiangsu (No. BF2024060). On computing resources, this work was supported by the Big Data Computing Center of Southeast University.

\vspace{-2mm}
\bibliographystyle{elsarticle-num}
\bibliography{cas-refs.bib}

\end{document}